%% file: draft.tex
\documentclass[10pt,twocolumn,letterpaper]{article}

\usepackage{cvpr}
\usepackage{times}
\usepackage{epsfig}
\usepackage{graphicx}
\usepackage{amsmath}
\usepackage{amssymb}
\usepackage{array}
\usepackage{enumitem}
\usepackage{multirow}
\usepackage[footnotesize]{subfigure}

\usepackage[lined,linesnumbered,ruled]{algorithm2e}

\SetCommentSty{mycommfont}

\usepackage[pagebackref=true,breaklinks=true,letterpaper=true,colorlinks,bookmarks=false]{hyperref}

\cvprfinalcopy 

\def\cvprPaperID{2402} 

\begin{document}

\title{SCA-CNN: Spatial and Channel-wise Attention in Convolutional Networks \\ for Image Captioning}
\author{Long Chen$^{1}$ \; Hanwang Zhang$^{2}$ \; Jun Xiao$^{1}$\thanks{Corresponding author} \; Liqiang Nie$^{3}$ \;
Jian Shao$^{1}$ \; Wei Liu$^{4}$ \; Tat-Seng Chua$^{5}$ \\
$^{1}$Zhejiang University \quad $^{2}$Columbia University \quad $^{3}$Shandong University \\
$^{4}$Tencent AI Lab \quad $^{5}$National University of Singapore \\
}


\maketitle
\thispagestyle{empty}

\begin{abstract}
Visual attention has been successfully applied in structural prediction tasks such as visual captioning and question answering. Existing visual attention
models are generally spatial, i.e., the attention is modeled as spatial probabilities that re-weight the last conv-layer feature map of a CNN encoding an
input image. However, we argue that such spatial attention does not necessarily conform to the attention mechanism --- a dynamic feature extractor that
combines contextual fixations over time, as CNN features are naturally spatial, channel-wise and multi-layer. In this paper, we introduce a novel
convolutional neural network dubbed SCA-CNN that incorporates Spatial and Channel-wise Attentions in a CNN. In the task of image captioning, SCA-CNN
dynamically modulates the sentence generation context in multi-layer feature maps, encoding where (i.e., attentive spatial locations at multiple layers) and
what (i.e., attentive channels) the visual attention is. We evaluate the proposed SCA-CNN architecture on three benchmark image captioning datasets:
Flickr8K, Flickr30K, and MSCOCO. It is consistently observed that SCA-CNN significantly outperforms state-of-the-art visual attention-based image captioning
 methods.
\end{abstract}
\input{intro}
\input{rew}

\input{app}
\input{exp}
\input{conc}


{ \footnotesize
\bibliographystyle{ieee}
\bibliography{egbib}
}

\end{document}

%% file: intro.tex
\section{Introduction}
Visual attention has been shown effective in various structural prediction tasks such as image/video captioning~\cite{xu2015show,yao2015describing} and visual
question answering~\cite{chen2015abc,yang2015stacked,xu2015ask}. Its success is mainly due to the reasonable assumption that human vision does not tend to
process a whole image in its entirety at once; instead, one only focuses on selective parts of the whole visual space when and where as
needed~\cite{corbetta2002control}. Specifically, rather than encoding an image into a static vector, attention allows the image feature to evolve from the
sentence context at hand, resulting in richer and longer descriptions for cluttered images. In this way, visual attention can be considered as a dynamic
feature extraction mechanism that combines contextual fixations over time~\cite{mnih2014recurrent,stollenga2014deep}.

\begin{figure}
	\centering
	\includegraphics[width=.9\linewidth]{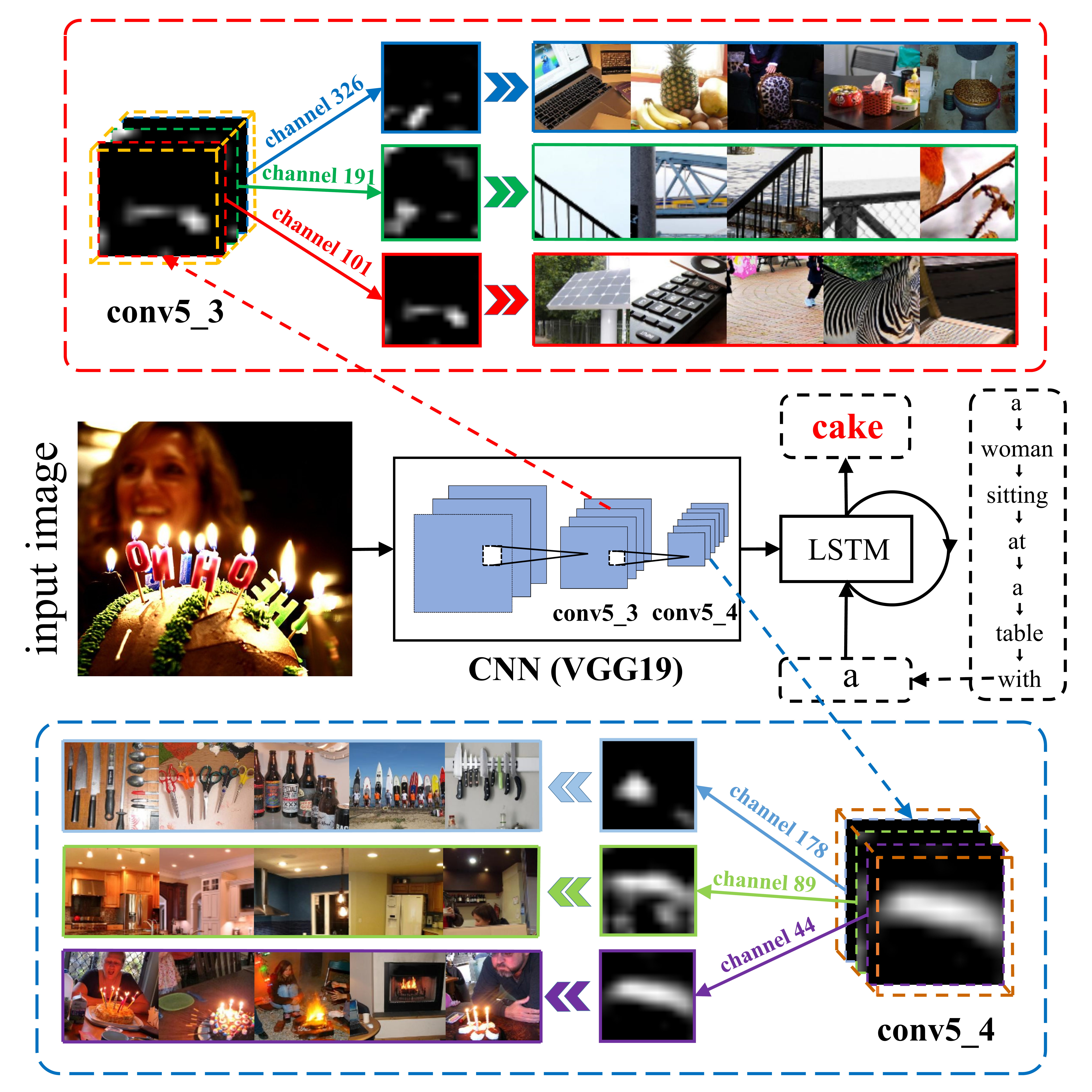}
	\caption{The illustration of channel-wise visual attention in two convolutional layers ($conv5\_3$ and $conv5\_4$ in VGG19) when predicting \texttt{cake} from the captioning \texttt{a woman sitting at a table with cake}. At each layer, top 3 attentive channels are visualized by showing the 5 most responsive receptive fields in the corresponding feature maps~\cite{zeiler2014visualizing}.}
	\label{fig:1}
\end{figure}

State-of-the-art image features are generally extracted by deep Convolutional Neural Networks (CNNs)~\cite{he2015deep,simonyan2014very,wei2015hcp}. Starting
from an input color image of the size $W\times H\times3$, a convolutional layer consisting of $C$-channel filters scans the input image and output a $W'\times
H'\times C$ feature map, which will be the input for the next convolutional layer\footnote[1]{Each convolutional layer is optionally followed by a pooling,
down-sampling, normalization, or a fully connected layer.}. Each 2D slice of a 3D feature map encodes the spatial visual responses raised by a filter channel,
where the filter performs as a pattern detector --- lower-layer filters detect low-level visual cues like edges and corners while higher-level ones detect
high-level semantic patterns like parts and object~\cite{zeiler2014visualizing}. By stacking the layers, a CNN extracts image features through a hierarchy of
visual abstractions. Therefore, CNN image features are essentially \textit{spatial}, \textit{channel-wise}, and \textit{multi-layer}. However, most existing
attention-based image captioning models only take into account the spatial characteristic~\cite{xu2015show}, \ie, those attention models merely modulate the
sentence context into the last conv-layer feature map via spatially attentive weights.

In this paper, we will take full advantage of the three characteristics of CNN features for visual attention-based image captioning. In particular, we propose
a novel Spatial and Channel-wise Attention-based Convolutional Neural Network, dubbed SCA-CNN, which learns to pay attention to every feature entry in the
multi-layer 3D feature maps. Figure~\ref{fig:1} illustrates the motivation of introducing channel-wise attention in multi-layer feature maps. First, since a
channel-wise feature map is essentially a detector response map of the corresponding filter, channel-wise attention can be viewed as the process of selecting
semantic attributes on the demand of the sentence context. For example, when we want to predict \texttt{cake}, our channel-wise attention (\eg, in the
$conv5\_3$/$conv5\_4$ feature map) will assign more weights on channel-wise feature maps generated by filters according to the semantics like cake, fire,
light, and candle-like shapes. Second, as a feature map is dependent on its lower-layer ones, it is natural to apply attention in multiple layers, so as to
gain visual attention on multiple semantic abstractions. For example, it is beneficial to emphasize on lower-layer channels corresponding to more elemental
shapes like array and cylinder that compose \texttt{cake}.

We validate the effectiveness of the proposed SCA-CNN on three well-known image captioning benchmarks: Flickr8K, Flickr30K and MSCOCO. SCA-CNN can
significantly surpass the spatial attention model~\cite{xu2015show} by $4.8\%$ in BLEU4. In summary, we propose a unified SCA-CNN framework to effectively
integrate spatial, channel-wise, and multi-layer visual attention in CNN features for image captioning. In particular, a novel spatial and channel-wise
attention model is proposed. This model is generic and thus can be applied to any layer in any CNN architecture such as popular VGG~\cite{simonyan2014very}
and ResNet~\cite{he2015deep}. SCA-CNN helps us gain a better understanding of how CNN features evolve in the process of the sentence generation.

%% file: rew.tex
\section{Related Work}

We are interested in visual attention models used in the encoder-decoder framework for neural image/video captioning (NIC) and visual question answering
(VQA), which fall into the recent trend of connecting computer vision and natural
language~\cite{krishna2016visual,zhang2017relation,shen2013inductive,shen2015supervised,zhao2016partial,jiang2015deep}. Pioneering work on
NIC~\cite{vinyals2015show,karpathy2015deep,donahue2015long,venugopalan2014translating,venugopalan2015sequence} and
VQA~\cite{antol2015vqa,malinowski2015ask,gao2015you,ren2015exploring} uses a CNN to encode an image or video into a static visual feature vector and then feed
it into an RNN~\cite{lstm} to decode language sequences such as captions or answers.

However, the static vector does not allow the image feature adapting to the sentence context at hand. Inspired by the attention mechanism introduced in
machine translation~\cite{bahdanau2014neural}, where a decoder dynamically selects useful source language words or sub-sequence for the translation into a
target language, visual attention models have been widely-used in NIC and VQA. We categorize these attention-based models into the following three domains
that motivate our SCA-CNN:

\leftmargini=3.5mm
\begin{itemize}
     \item{\textbf{Spatial Attention}}. Xu~\etal~\cite{xu2015show} proposed the first visual attention model in image captioning. In general, they used
         ``hard'' pooling that selects the most probably attentive region, or ``soft'' pooling that averages the spatial features with attentive weights. As
         for VQA, Zhu~\etal~\cite{zhu2015visual7w} adopted the ``soft'' attention to merge image region features. To further refine the spatial attention,
         Yang~\etal~\cite{yang2015stacked} and Xu~\etal~\cite{xu2015ask} applied a stacked spatial attention model, where the second attention is based on
         the attentive feature map modulated by the first one. Different from theirs, our multi-layer attention is applied on the multiple layers of a CNN.
         A common defect of the above spatial models is that they generally resort to weighted pooling on the attentive feature map. Thus, spatial
         information will be lost inevitably. More seriously, their attention is only applied in the last conv-layer, where the size of receptive field will
         be quite large and the differences between each receptive field region are quite limited, resulting in insignificant spatial attentions.

    \item{\textbf{Semantic Attention}}. Besides the spatial information, You~\etal~\cite{you2016image} proposed to select semantic concepts in NIC, where
        the image feature is a vector of confidences of attribute classifiers. Jia~\etal~\cite{jia2015guiding} exploited the correlation between images and
        their captions as the global semantic information to guide the LSTM generating sentences. However, these models require external resources to train
        these semantic attributes. In SCA-CNN, each filter kernel of a convolutional layer servers as a semantic detectors~\cite{zeiler2014visualizing}.
        Therefore, the channel-wise attention of SCA-CNN is similar to semantic attention.

    \item{\textbf{Multi-layer Attention}}. According to the nature of CNN architecture, the sizes of respective fields corresponding to different feature
        map layers are different. To overcome the weakness of large respective field size in the last conv-layer attention,
        Seo~\etal~\cite{seo2016hierarchical} proposed a multi-layer attention networks. In compared with theirs, SCA-CNN also incorporates the channel-wise
        attention at multiple layers.

\end{itemize}

%% file: app.tex
\section{Spatial and Channel-wise Attention CNN}

\begin{figure*}[t]
	\centering
	\includegraphics[width=.9\textwidth]{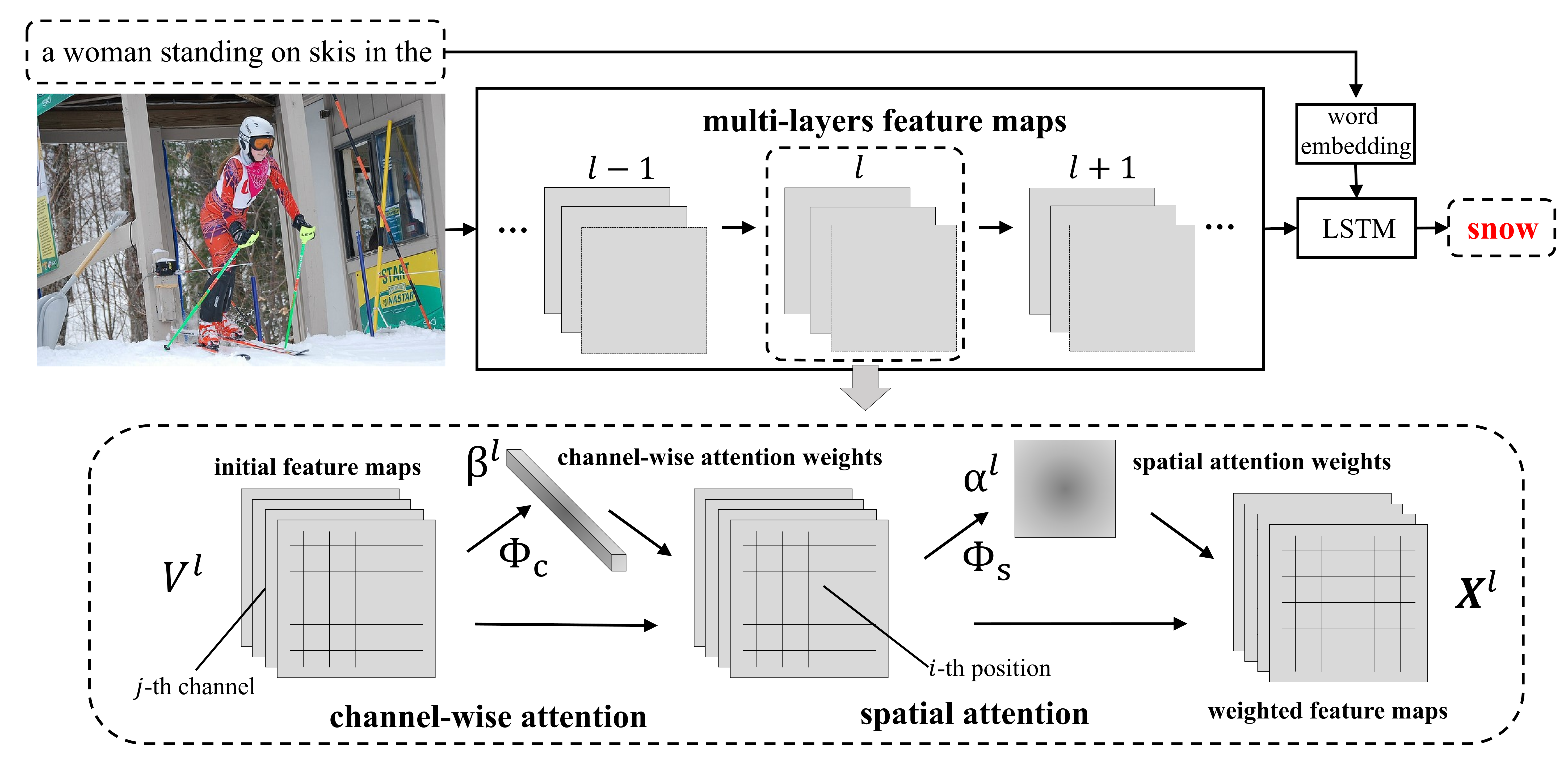}
    \caption{The overview of our proposed SCA-CNN. For the $l$-th layer, initial feature map $\mathbf{V}^l$ is the output of $(l-1)$-th conv-layer.
    We first use the channel-wise attention function $\Phi_c$ to obtain the channel-wise attention weights $\beta^l$, which are multiplied in channel-wise of the
    feature map. Then, we use the spatial attention function $\Phi_s$ to obtain the spatial attention weights $\alpha^l$, which are multiplied in each spatial
    regions, resulting in an attentive feature map $\mathbf{X}^l$. Different orders of two attention mechanism are discussed in Section~\ref{sec:channel}.}
\label{fig:2}
\end{figure*}

\subsection{Overview}
We adopt the popular encoder-decoder framework for image caption generation, where a CNN first encodes an input image into a vector and then an LSTM decodes
the vector into a sequence of words. As illustrated in Figure~\ref{fig:2}, SCA-CNN makes the original CNN multi-layer feature maps adaptive to the sentence
context through channel-wise attention and spatial attention at multiple layers.

Formally, suppose that we want to generate the $t$-th word of the image caption. At hand, we have the last sentence context encoded in the LSTM memory
$\mathbf{h}_{t-1}\in\mathbb{R}^d$, where $d$ is the hidden state dimension. At the $l$-th layer, the spatial and channel-wise attention weights $\gamma^l$ are
a function of $\mathbf{h}_{t-1}$ and the current CNN features $\mathbf{V}^l$. Thus, SCA-CNN modulates $\mathbf{V}^l$ using the attention weights $\gamma^l$ in
a recurrent and multi-layer fashion as:
\begin{equation} \label{equ:whole_attention}
\begin{split}
\mathbf{V}^l &= \textrm{CNN}\left(\mathbf{X}^{l-1}\right),\\
\gamma^l &= \Phi\left(\mathbf{h}_{t-1},\mathbf{V}^l\right),\\
\mathbf{X}^l &= f\left(\mathbf{V}^{l},\gamma^{l}\right).
\end{split}
\end{equation}
where $\mathbf{X}^l$ is the modulated feature, $\Phi(\cdot)$ is the spatial and channel-wise attention function that will be detailed in
Section~\ref{sec:spatial} and~\ref{sec:channel}, $\mathbf{V}^l$ is the feature map output from previous conv-layer, \eg, convolution followed by pooling,
down-sampling or convolution~\cite{simonyan2014very,he2015deep}, and $f(\cdot)$ is a linear weighting function that modulates CNN features and attention
weights. Different from existing popular modulating strategy that sums up all visual features based on attention weights~\cite{xu2015show}, function
$f(\cdot)$ applies element-wise multiplication.
So far, we are ready to generate the $t$-th word by:
\begin{equation}
\begin{split}
\mathbf{h}_t &= \textrm{LSTM}\left(\mathbf{h}_{t-1},\mathbf{X}^L,y_{t-1}\right),\\
y_t & \sim p_t = \textrm{softmax} \left(\mathbf{h}_t, y_{t-1} \right).
\end{split}
\end{equation}
where $L$  is the total number of conv-layers; $p_t \in \mathbb{R}^{|\mathcal{D}|}$ is a probability vector and $\mathcal{D}$ is a predefined dictionary
including all caption words.

Note that $\gamma^l$ is of the same size as $\mathbf{V}^l$ or $\mathbf{X}^l$, \ie, $W^l\times H^l\times C^l$. It will require $\mathcal{O}(W^lH^lC^lk)$ space
for attention computation, where $k$ is the common mapping space dimension of CNN feature $\mathbf{V}^l$ and hidden state $\mathbf{h}_{t-1}$. It is
prohibitively expensive for GPU memory when the feature map size is so large. Therefore, we propose an approximation that learns spatial attention weights
$\alpha^l$ and channel-wise attention weights $\beta^l$ separately:
\begin{eqnarray}
\alpha^l &= & \Phi_s \left(\mathbf{h}_{t-1},\mathbf{V}^l\right),  \label{equ:spatial} \\
\beta^l &= & \Phi_c \left(\mathbf{h}_{t-1},\mathbf{V}^l\right). \label{equ:channel}
\end{eqnarray}
Where $\Phi_c$ and $\Phi_s$ represent channel-wise and spatial attention model respectively. This will greatly reduce the memory cost into
$\mathcal{O}(W^lH^lk)$ for spatial attention and $\mathcal{O}(C^lk)$ for channel-wise attention, respectively.

\subsection{Spatial Attention}\label{sec:spatial}
In general, a caption word only relates to partial regions of an image. For example, in Figure~\ref{fig:1}, when we want to predict \texttt{cake}, only image
regions which contain cake are useful. Therefore, applying a global image feature vector to generate caption may lead to sub-optimal results due to the
irrelevant regions. Instead of considering each image region equally, spatial attention mechanism attempts to pay more attention to the semantic-related
regions. Without loss of generality, we discard the layer-wise superscript $l$. We reshape $\mathbf{V}  = \left[\mathbf{v}_1, \mathbf{v}_2, ..., \mathbf{v}_m
\right]$ by flattening the width and height of the original $\mathbf{V}$, where $\mathbf{v}_i\in\mathbb{R}^C$ and $m=W\cdot H$. We can consider $\mathbf{v}_i$
as the visual feature of the $i$-th location. Given the previous time step LSTM hidden state $\mathbf{h}_{t-1}$, we use a single-layer neural network followed
by a softmax function to generate the attention distributions $\alpha$ over the image regions. Below are the definitions of the spatial attention model
$\Phi_s$:
\begin{equation} \label{equ:S}
\begin{split}
\mathbf{a} & = \tanh \left( \left( \mathbf{W}_s \mathbf{V} + b_s \right) \oplus \mathbf{W}_{hs}\mathbf{h}_{t-1}\right), \\
\alpha & = \textrm{softmax} \left( \mathbf{W}_i \mathbf{a} + b_i \right).
\end{split}
\end{equation}
where $\mathbf{W}_s \in \mathbb{R}^{k \times C}, \mathbf{W}_{hs} \in \mathbb{R}^{k \times d}, \mathbf{W}_i \in \mathbb{R}^k$ are transformation matrices that
map image visual features and hidden state to a same dimension. We denote $\oplus$ as the addition of a matrix and a vector. And the addition between a matrix
and a vector is performed by adding each column of the matrix by the vector. $b_s \in \mathbb{R}^k, b_i \in \mathbb{R}^1$ are model biases.

\subsection{Channel-wise Attention}\label{sec:channel}
Note that the spatial attention function in Eq~\eqref{equ:spatial} still requires the visual feature $\mathbf{V}$ to calculate the spatial attention weights,
but the visual feature $\mathbf{V}$ used in spatial attention is in fact not attention-based. Hence, we introduce a channel-wise attention mechanism to attend
the features $\mathbf{V}$. It is worth noting that each CNN filter performs as a pattern detector, and each channel of a feature map in CNN is a response
activation of the corresponding convolutional filter. Therefore, applying an attention mechanism in channel-wise manner can be viewed as a process of
selecting semantic attributes.

For channel-wise attention, we first reshape $\mathbf{V}$ to $\mathbf{U}$, and $ \mathbf{U} = [\mathbf{u}_1, \mathbf{u}_2, ..., \mathbf{u}_C]$, where $
\mathbf{u}_i \in \mathbb{R}^{W \times H}$ represents the $i$-th channel of the feature map $\mathbf{V}$, and $C$ is the total number of channels. Then, we
apply mean pooling for each channel to obtain the channel feature $\mathbf{v}$:
\begin{equation}
\mathbf{v} = \left[v_1, v_2, ..., v_C \right], \mathbf{v} \in \mathbb{R}^{C},
\end{equation}
where scalar $v_i$ is the mean of vector $\mathbf{u}_i$, which represents the $i$-th channel features. Following the definition of the spatial attention
model, the channel-wise attention model $\Phi_c$ can be defined as follows:
\begin{equation} \label{equ:C}
\begin{split}
\mathbf{b} & = \tanh \left(\left(\mathbf{W}_c \otimes \mathbf{v} + b_c \right) \oplus \mathbf{W}_{hc}\mathbf{h}_{t-1} \right), \\
\beta & = \textrm{softmax} \left(\mathbf{W'}_i \mathbf{b} + {b'}_i \right).
\end{split}
\end{equation}
where $\mathbf{W}_c \in \mathbb{R}^k, \mathbf{W}_{hc} \in \mathbb{R}^{k \times d}, \mathbf{W'}_i \in \mathbb{R}^k$ are transformation matrices, $\otimes$
represents the outer product of vectors. $b_c \in \mathbb{R}^k, {b'}_i \in \mathbb{R}^1$ are bias terms.

According to different implementation order of channel-wise attention and spatial attention, there exists two types of model which incorporating both two
attention mechanisms. We distinguish between the two types as follows:

\textbf{Channel-Spatial}. The first type dubbed Channel-Spatial (C-S) applies channel-wise attention before spatial attention. The flow chart of C-S type is
illustrated in Figure~\ref{fig:2}. At first, given an initial feature map $\mathbf{V}$, we adopt channel-wise attention $\Phi_c$ to obtain the channel-wise
attention weights $\beta$. Through a linear combination of $\beta$ and $\mathbf{V}$, we obtain a channel-wise weighted feature map. Then we feed the
channel-wise weighted feature map to the spatial attention model $\Phi_s$ and obtain the spatial attention weights $\alpha$. After attaining two attention
weights $\alpha$ and $\beta$, we can feed $\mathbf{V}, \beta, \alpha$ to modulate function $f$ to calculate the modulated feature map $\mathbf{X}$. All
processes are summarized as follows:
\begin{equation} \label{equ:C-S}
\begin{split}
\beta &= \Phi_c \left(\mathbf{h}_{t-1},\mathbf{V} \right), \\
\alpha &= \Phi_s \left(\mathbf{h}_{t-1}, f_c \left(\mathbf{V}, \beta \right) \right), \\
\mathbf{X} &= f \left(\mathbf{V}, \alpha, \beta \right).
\end{split}
\end{equation}
where $f_c(\cdot)$ is a channel-wise multiplication for feature map channels and corresponding channel weights.

\textbf{Spatial-Channel}. The second type denoted as Spatial-Channel (S-C) is a model with spatial attention implemented first. For S-C type, given an initial
feature map $\mathbf{V}$, we first utilize spatial attention $\Phi_s$ to obtain the spatial attention weights $\alpha$. Based on $\alpha$, the linear function
$f_s(\cdot)$, and the channel-wise attention model $\Phi_c$, we can calculate the modulated feature $\mathbf{X}$ following the recipe of C-S type:
\begin{equation} \label{equ:S-C}
\begin{split}
\alpha &= \Phi_s \left(\mathbf{h}_{t-1}, \mathbf{V} \right), \\
\beta &= \Phi_c \left(\mathbf{h}_{t-1}, f_s \left(\mathbf{V}, \alpha \right) \right), \\
\mathbf{X} &= f \left(\mathbf{V}, \alpha, \beta \right).
\end{split}
\end{equation}
where $f_s(\cdot)$ is an element-wise multiplication for regions of each feature map channel and its corresponding region attention weights.

%% file: exp.tex
\section{Experiments}
We will validate the effectiveness of the proposed SCA-CNN framework for image captioning by answering the following questions:
\textbf{Q1} Is the channel-wise attention effective? Will it improve the spatial attention? \textbf{Q2} Is the multi-layer attention effective? \textbf{Q3}
How does SCA-CNN perform compared to other state-of-the-art visual attention models?

\subsection{Dataset and Metric}
We conducted experiments on three well-known benchmarks: 1) \textbf{Flickr8k}~\cite{hodosh2013framing}: it contains 8,000 images. According to its official
split, it selects 6,000 images for training, 1,000 images for validation, and 1,000 images for testing; 2) \textbf{Flickr30k}~\cite{young2014image}: it
contains 31,000 images. Because of the lack of official split, for fair comparison with previous works, we reported results in a publicly available split used
in previous work~\cite{karpathy2015deep}. In this split, 29,000 images are used for training, 1,000 images for validation, and 1,000 images for testing; and
3) \textbf{MSCOCO}~\cite{lin2014microsoft}: it contains 82,783 images in training set, 40,504 images in validation set and 40,775 images in test set. As the
ground truth of MSCOCO test set is not available, the validation set is further splited into a validation subset for model selection and a test subset for
local experiments. This split also follows~\cite{karpathy2015deep}. It utilizes the whole 82,783 training set images for training, and selects 5,000 images
for validation and 5,000 images for test from official validation set .
As for the sentences preprocessing, we followed the publicly available code~\footnote{https://github.com/karpathy/neuraltalk}.
We used BLEU (\textbf{B@1},\textbf{B@2}, \textbf{B@3}, \textbf{B@4})~\cite{papineni2002bleu}, METEOR (\textbf{MT})~\cite{banerjee2005meteor},
CIDEr(\textbf{CD})~\cite{vedantam2015cider}, and ROUGE-L (\textbf{RG})~\cite{lin2004rouge} as evaluation metrics. For all the four metrics, in a nutshell,
they measure the consistency between n-gram occurrences in generated sentences and ground-truth sentences, where this consistency is weighted by n-gram
saliency and rarity. Meanwhile, all the four metrics can be calculated directly through the MSCOCO caption evaluation
tool\footnote{https://github.com/tylin/coco-caption}. And our source code is already publicly available~\footnote{https://github.com/zjuchenlong/sca-cnn}.

\subsection{Setup}
In our captioning system, for image encoding part, we adopted two widely-used CNN architectures: VGG-19~\cite{simonyan2014very} and
ResNet-152~\cite{he2015deep} as the basic CNNs for SCA-CNN. For the caption decoding part, we used an LSTM~\cite{lstm} to generate caption words. Word
embedding dimension and LSTM hidden state dimension are respectively set to 100 and 1,000. The common space dimension for calculating attention weights is set
to 512 for both two type attention.
For Flickr8k, mini-batch size is set to 16, and for Flickr30k and MSCOCO, mini-batch size is set to 64. We use dropout and early stopping to avoid
overfitting. Our whole framework is trained in an end-to-end way with Adadelta~\cite{adadelta}, which is a stochastic gradient descent method using an
adaptive learning rate algorithm. The caption generation process would be halted until a special END token is predicted or a predefined max sentence length is
reached. We followed the strategy of BeamSearch~\cite{vinyals2015show} in the testing period, which selects the best caption from some candidates, and the
beam size is set to 5. We noticed a trick that incorporates beam search with length normalization~\cite{jia2015guiding} which can help to improve performance
in some degree. But for fair comparisons, all results reported are without length normalization.

\subsection{Evaluations of Channel-wise Attention (Q1)} \label{sec:Q1}

\textbf{Comparing Methods}. We first compared spatial attention with channel-wise attention. 1) \textbf{S}: It is a pure spatial attention model. After
obtaining spatial attention weights based on the last conv-layer, we use element-wise multiplication to produce a spatial weighted feature. For VGG-19 and
ResNet-152, the last conv-layer represents $conv5\_4$ layer and $res5c$, respectively. Instead of regarding the weighted feature map as the final visual
representation, we feed the spatial weighted feature into their own following CNN layers. For VGG-19, there are two fully-connected layers follows $conv5\_4$
layer and for ResNet-152, $res5c$ layer is followed by a mean pooling layer. 2) \textbf{C}: It is a pure channel-wise attention model. The whole strategy for
the C type model is same as S type. The only difference is substituting the spatial attention with channel-wise attention as Eq.~\eqref{equ:channel}. 3)
\textbf{C-S}: This is the first type model incorporating two attention mechanisms as Eq.~\eqref{equ:C-S}. 4) \textbf{S-C}: Another incorporating model
introduced in Eq.~\eqref{equ:S-C}. 5) \textbf{SAT}: It is the ``hard" attention model introduced in~\cite{xu2015show}. The reason why we report the results of
``hard" attention instead of the ``soft" attention is that ``hard" attention always has better performance on different datasets and metrics. SAT is also a
pure spatial attention model like S. But there are two main differences. The first one is the strategy of modulating visual feature with attention weights.
The second one is whether to feed the attending features into their following layers. All VGG results reported in Table~\ref{tab:Q1} came from the original
paper and ResNet results are our own implementation.

\begin{table}[htbp]
\small
\centering
\scalebox{0.8}{
\begin{tabular}{l| l |l| c c c c}
\hline
 Dataset & Network &Method & B@4 & MT & RG & CD\\
\hline
\multirow{10}{*}{Flickr8k} & \multirow{5}{*}{VGG} & S & 23.0 & 21.0 & 49.1 & \textbf{60.6} \\
&  & SAT & 21.3 & 20.3 & --- & --- \\
&  & C & 22.6 & 20.3 & 48.7 & 58.7 \\
& & S-C &22.6 & 20.9 & 48.7 & \textbf{60.6}\\
& & C-S & \textbf{23.5} & \textbf{21.1} & \textbf{49.2} & 60.3\\
\cline{2-7}
& \multirow{5}{*}{ResNet} & S & 20.5 &　19.6 & 47.4 & 49.9 \\
&  & SAT & 21.7 & 20.1 & 48.4 & 55.5 \\
&  & C & 24.4 & 21.5 & 50.0 & 65.5 \\
& & S-C &24.8 & \textbf{22.2} & 50.5& 65.1\\
& & C-S & \textbf{25.7} & 22.1& \textbf{50.9} & \textbf{66.5}\\
\hline
\multirow{10}{*}{Flickr30k} & \multirow{5}{*}{VGG} & S & \textbf{21.1} & 18.4 & 43.1 & \textbf{39.5} \\
&  & SAT & 19.9 & \textbf{18.5} &  --- & --- \\
&  & C & 20.1 & 18.0 & 42.7 & 38.0 \\
& & S-C & 20.8 & 17.8&42.9 & 38.2\\
& & C-S & 21.0 & 18.0& \textbf{43.3} & 38.5\\
\cline{2-7}
& \multirow{5}{*}{ResNet} & S & 20.5 &　17.4 & 42.8 & 35.3 \\
&  & SAT & 20.1 & 17.8 & 42.9 & 36.3 \\
&  & C & 21.5 & 18.4 & 43.8 & 42.2 \\
& & S-C &21.9 & 18.5 & 44.0& \textbf{43.1}\\
& & C-S & \textbf{22.1} & \textbf{19.0} & \textbf{44.6} & 42.5 \\
\hline
\multirow{10}{*}{MS COCO} & \multirow{5}{*}{VGG} & S & \textbf{28.2} & 23.3 & \textbf{51.0} & \textbf{85.7} \\
&  & SAT & 25.0 & 23.0 & --- & --- \\
&  & C & 27.3 & 22.7 & 50.1 & 83.4 \\
& & S-C &28.0 &23.0 & 50.6& 84.9\\
& & C-S & 28.1 & \textbf{23.5} & 50.9& 84.7\\
\cline{2-7}
& \multirow{5}{*}{ResNet} & S & 28.3 &　23.1 & 51.2 & 84.0 \\
&  & SAT & 28.4 & 23.2 & 51.2 & 84.9 \\
&  & C & 29.5 & 23.7 & 51.8 & 91.0 \\
& & S-C &29.8 &23.9 & 52.0& 91.2\\
& & C-S & \textbf{30.4} & \textbf{24.5} & \textbf{52.5} & \textbf{91.7}\\
\hline
\end{tabular}}
\vspace{2pt}
\caption{The performance of S, C, C-S, S-C, SAT with one attentive layer in VGG-19 and ResNet-152.} \label{tab:Q1}
\end{table}

\textbf{Results} From Table~\ref{tab:Q1}, we have the following observations: 1) For VGG-19, performance of S is better than that of SAT; but for ResNet-152,
the results are opposite. This is because the VGG-19 network has fully-connected layers, which can preserve spatial information. Instead, in ResNet-152, the
last conv-layer is originally followed by an average pooling layer, which can destroy spatial information. 2) Comparing to the performance of S, the
performance of C can be significant improved in ResNet-152 rather than VGG-19. It shows that the more channel numbers can help improve channel-wise attention
performance in the sense that ResNet-152 has more channel numbers (\ie2048) than VGG-19 (\ie512). 3) In ResNet-152, both C-S and S-C can achieve better
performance than S. This demonstrates that we can improve performance significantly by adding channel-wise attention as long as channel numbers are large. 4)
In both of two networks, the performance of S-C and C-S is quite close. Generally, C-S is slightly better than S-C, so in the following experiments we use C-S
to represent incorporating model.

\subsection{Evaluations of Multi-layer Attention (Q2)} \label{sec:Q2}

\begin{table}[htbp]
\small
\centering
\scalebox{0.8}{
\begin{tabular}{l| l |l| c c c c}
\hline
 Dataset & Network &Method & B@4 & MT & RG & CD\\
\hline
\multirow{6}{*}{Flickr8k} & \multirow{3}{*}{VGG} & 1-layer & \textbf{23.0} & 21.0 & \textbf{49.1} & \textbf{60.6} \\
&  & 2-layer & 22.8 & \textbf{21.2} & 49.0 & 60.4 \\
&  & 3-layer & 21.6 & 20.9 & 48.4 & 54.5 \\
\cline{2-7}
& \multirow{3}{*}{ResNet} & 1-layer & 20.5 &　19.6 & 47.4 & 49.9 \\
&  & 2-layer & 22.9 & 21.2 & 48.8 & 58.8 \\
&  & 3-layer & \textbf{23.9} & \textbf{21.3} & \textbf{49.7} & \textbf{61.7} \\
\hline
\multirow{6}{*}{Flickr30k} & \multirow{3}{*}{VGG} & 1-layer & 21.1 & 18.4 & 43.1 & \textbf{39.5} \\
&  & 2-layer & \textbf{21.9} & \textbf{18.5} & \textbf{44.3} & \textbf{39.5} \\
&  & 3-layer & 20.8 & 18.0 & 43.0 & 38.5 \\  
\cline{2-7}
& \multirow{3}{*}{ResNet} & 1-layer & 20.5 &　17.4 & 42.8 & 35.3 \\
&  & 2-layer & 20.6 & 18.6 & 43.2 & 39.7 \\
&  & 3-layer & \textbf{21.0} & \textbf{19.2} & \textbf{43.4} & \textbf{43.5} \\
\hline
\multirow{6}{*}{MS COCO} & \multirow{3}{*}{VGG} & 1-layer & 28.2 & 23.3 & 51.0 & 85.7 \\
&  & 2-layer & \textbf{29.0} & \textbf{23.6} & \textbf{51.4} & \textbf{87.4}\\
&  & 3-layer & 27.4 & 22.9 & 50.4 & 80.8 \\
\cline{2-7}
& \multirow{3}{*}{ResNet} & 1-layer & 28.3 &　23.1 & 51.2 & 84.0 \\
&  & 2-layer & \textbf{29.7} & 24.1 & \textbf{52.2} & \textbf{91.1} \\
&  & 3-layer & 29.6 & \textbf{24.2} & 52.1 & 90.3 \\
\hline
\end{tabular}}
\vspace{5pt}
\caption{The performance of multi-layer in S in both VGG-19 network and ResNet-152 network} \label{tab:Q2_1}
\end{table}

\begin{table}[htbp]
\small
\centering
\scalebox{0.8}{
\begin{tabular}{l| l |l| c c c c}
\hline
 Dataset & Network &Method & B@4 & MT & RG & CD\\
\hline
\multirow{6}{*}{Flickr8k} & \multirow{3}{*}{VGG} & 1-layer & \textbf{23.5} & 21.1 & 49.2 & 60.3 \\
&  & 2-layers & 22.8 & \textbf{21.6} & \textbf{49.5} & 62.1 \\
&  & 3-layers & 22.7 & 21.3 & 49.3 & \textbf{62.3} \\
\cline{2-7}
& \multirow{3}{*}{ResNet} & 1-layer & 25.7 &　22.1 & 50.9 & 66.5 \\
&  & 2-layers & \textbf{25.8} & 22.4 & \textbf{51.3} & 67.1 \\
&  & 3-layers & 25.3 & \textbf{22.9} & 51.2 & \textbf{67.5} \\
\hline
\multirow{6}{*}{Flickr30k} & \multirow{3}{*}{VGG} & 1-layer & 21.0 & 18.0 & 43.3 & 38.5 \\
&  & 2-layers & \textbf{21.8} & \textbf{18.8} & \textbf{43.7} & \textbf{41.4} \\
&  & 3-layers & 20.7 & 18.3 & 43.6 & 39.2 \\
\cline{2-7}
& \multirow{3}{*}{ResNet} & 1-layer & 22.1 &　19.0 & 44.6 & 42.5 \\
&  & 2-layers & \textbf{22.3} & \textbf{19.5} & \textbf{44.9} & \textbf{44.7} \\
&  & 3-layers & 22.0 & 19.2 & 44.7 & 42.8 \\        
\hline
\multirow{6}{*}{MS COCO} & \multirow{3}{*}{VGG} & 1-layer & 28.1 & 23.5 & 50.9 & 84.7 \\
&  & 2-layers & \textbf{29.8} & \textbf{24.2} & \textbf{51.9} & \textbf{89.7} \\
&  & 3-layers & 29.4 & 24.0 & 51.7 & 88.4 \\    
\cline{2-7}
& \multirow{3}{*}{ResNet} & 1-layer & 30.4 &　24.5 & 52.5 & 91.7 \\
&  & 2-layers & \textbf{31.1} & \textbf{25.0} & \textbf{53.1} & \textbf{95.2} \\
&  & 3-layers & 30.9 & 24.8 & 53.0 & 94.7 \\   
\hline
\end{tabular}}
\vspace{2pt}
\caption{The performance of multi-layer in C-S in both VGG-19 network and ResNet-152 network} \label{tab:Q2_2}
\end{table}

\begin{table*}[htbp]
\centering
\scalebox{0.85}{
\begin{tabular}{ l | c c c c c  | c c c c c | c c c c c }
\hline
\multirow{2}{*}{Model} & \multicolumn{5}{c|}{Flickr8k} & \multicolumn{5}{c|}{Flickr30k} & \multicolumn{5}{c}{MS COCO} \\
\cline{2-16}
& B@1 & B@2 & B@3 & B@4 & MT & B@1 & B@2 & B@3 & B@4 & MT & B@1 & B@2 & B@3 & B@4 & MT \\
\hline
Deep VS~\cite{karpathy2015deep} & 57.9 & 38.3 & 24.5 & 16.0 &　-- &  57.3 & 36.9 & 24.0 & 15.7 & --  & 62.5 & 45.0 & 32.1 & 23.0 &　19.5 \\
Google NIC~\cite{vinyals2015show}${}^\dagger$ & 63.0 & 41.0 & 27.0 & -- &　-- & 66.3 & 42.3 & 27.7 & 18.3  & -- & 66.6 & 46.1 & 32.9 & 24.6 & 　-- \\
m-RNN~\cite{mao2014deep} & -- & -- & -- & -- &　-- & 60.0 & 41.0 & 28.0 & 19.0 & -- & 67.0 & 49.0 & 35.0 & 25.0 &　-- \\
Soft-Attention~\cite{xu2015show} & 67.0 & 44.8 & 29.9 & 19.5 &　18.9 & 66.7 & 43.4 & 28.8 & 19.1 & 18.5 & 70.7 & 49.2 & 34.4 & 24.3 &　23.9 \\
Hard-Attention~\cite{xu2015show} & 67.0 & 45.7 & 31.4 & 21.3 &　20.3 & \textbf{66.9} & 43.9 & 29.6 & 19.9 & 18.5 & 71.8 & 50.4 & 35.7 & 25.0 & 　23.0 \\
emb-gLSTM~\cite{jia2015guiding} & 64.7 & 45.9 & 31.8 & 21.2 &　20.6 & 64.6 & 44.6 & 30.5 & 20.6 & 17.9 &  67.0 & 49.1 & 35.8 & 26.4 &　22.7 \\
ATT~\cite{you2016image}${}^\dagger$ & -- & -- & -- & -- &　--  & 64.7 & 46.0 & 32.4 & \textbf{23.0} & 18.9 & 70.9 & 53.7 & 40.2 & 30.4 &　24.3  \\
\hline
SCA-CNN-VGG  & 65.5 & 46.6 & 32.6 & 22.8 &　21.6 & 64.6 & 45.3 & 31.7 & 21.8 & 18.8 & 70.5 & 53.3 & 39.7 & 29.8 & 24.2\\
SCA-CNN-ResNet  & \textbf{68.2} & \textbf{49.6} & \textbf{35.9} & \textbf{25.8} &　\textbf{22.4} & 66.2 & \textbf{46.8} & \textbf{32.5} & 22.3 & \textbf{19.5} & \textbf{71.9} & \textbf{54.8} & \textbf{41.1} & \textbf{31.1} & \textbf{25.0}\\
\hline
\end{tabular}}
\vspace{2pt}
\caption{Performances compared with the state-of-art in Flickr8k, Flickr30k and MSCOCO dataset. SCA-CNN-VGG is our C-S 2-layer model based on VGG-19 network, and SCA-CNN-ResNet is our C-S 2-layer model based on ResNet-152 network. ${}^\dagger$ indicates an ensemble model results. (--) indicates an unknow metric} \label{tab:state}
\end{table*}

\begin{table*}[htbp]
\centering
\scalebox{0.85}{
\begin{tabular}{ l | c | c | c| c | c |c | c |c | c |c | c| c | c |c  }
\hline
\multirow{2}{*}{Model} & \multicolumn{2}{c|}{B@1} & \multicolumn{2}{c|}{B@2} & \multicolumn{2}{c|}{B@3} & \multicolumn{2}{c|}{B@4} &\multicolumn{2}{c|}{METEOR} & \multicolumn{2}{c|}{ROUGE-L} & \multicolumn{2}{c}{CIDEr} \\
\cline{2-15}
& c5 & c40 & c5 & c40 & c5 & c40 & c5 & c40 & c5 & c40 & c5 & c40 & c5 & c40 \\
\hline
SCA-CNN &71.2 & 89.4& 54.2& 80.2& 40.4& 69.1& 30.2& 57.9& 24.4 &　33.1& 52.4& 67.4& 91.2 & 92.1 \\
\hline
Hard-Attention  & 70.5& 88.1& 52.8& 77.9& 38.3 & 65.8 & 27.7 & 53.7 & 24.1 & 32.2& 51.6 &65.4 & 86.5 & 89.3\\
\hline
ATT${}^\dagger$ & \textbf{73.1} & \textbf{90.0} & \textbf{56.5} & \textbf{81.5} & \textbf{42.4} & \textbf{70.9} & \textbf{31.6} & \textbf{59.9} &25.0 &　\textbf{33.5} & 53.5& \textbf{68.2} & \textbf{95.3} & \textbf{95.8}   \\
\hline
Google NIC${}^\dagger$ & 71.3 & 89.5 & 54.2& 80.2& 40.7& 69.4& 30.9& 58.7 & \textbf{25.4} & \textbf{34.6} & 53.0 & \textbf{68.2} & 94.3 & 94.6 \\
\hline
\end{tabular}}
\vspace{2pt}
\caption{Performances of the proposed attention model on the onlines MSCOCO testing server. ${}^\dagger$ indicates an ensemble model results.} \label{tab:server}
\end{table*}

\textbf{Comparing Methods} We will investigate whether we can improve the spatial attention or channel-wise attention performance by adding more attentive
layers. We conduct ablation experiments about different number of attentive layer in S and C-S models. In particular, we denote \textbf{1-layer},
\textbf{2-layer}, \textbf{3-layer} as the number of layers equipped with attention, respectively. For VGG-19, 1-st layer, 2-nd layer, 3-rd layer represent
$conv5\_4, conv5\_3, conv5\_2$ conv-layer, respectively. As for ResNet-152, it represents $res5c, res5c\_branch2b, res5c\_branch2a$ conv-layer. Specifically,
our strategy for training more attentive layers model is to utilize previous trained attentive layer weights as initialization, which can significantly reduce
the training time and achieve better results than randomly initialized.

\textbf{Results} From Table~\ref{tab:Q2_1} and~\ref{tab:Q2_2}, we have following observations: 1) In most experiments, adding more attentive layers can
achieve better results among two models. The reason is that applying an attention mechanism in multi-layer can help gain visual attention on multiple level
semantic abstractions. 2) Too many layers are also prone to resulting in severe overfitting. For example, Flickr8k's performance is easier to degrade than
MSCOCO when adding more attentive layers, as the size of train set of Flickr8k (\ie 6,000) is much smaller than that of MSCOCO (\ie 82,783).

\subsection{Comparison with State-of-The-Arts (Q3)}

\textbf{Comparing Methods} We compared the proposed SCA-CNN with state-of-the-art image captioning models. 1) \textbf{Deep VS}~\cite{karpathy2015deep},
\textbf{m-RNN}~\cite{mao2014deep}, and \textbf{Google NIC}~\cite{vinyals2015show} are all end-to-end multimodal networks, which combine CNNs for image
encoding and RNN for sequence modeling. 2) \textbf{Soft-Attention}~\cite{xu2015show} and \textbf{Hard-Attention}~\cite{xu2015show} are both pure spatial
attention model. The ``soft" attention weighted sums up the visual features as the attending feature, while the ``hard" one randomly samples the region
feature as the attending feature. 3) \textbf{emb-gLSTM}~\cite{jia2015guiding} and \textbf{ATT}~\cite{you2016image} are both semantic attention models. For
emb-gLSTM, it utilizes correlation between image and its description as gloabl semantic information, and for ATT it utilizes visual concepts corresponded
words as semantic information. The results reported in Table~\ref{tab:state} are from the 2-layer C-S model for both VGG-19 and ResNet-152 network, since this
type model always obtains the best performance in previous experiments. Besides the three benchmarks, we also evaluated our model on MSCOCO Image Challenge
set c5 and c40 by uploading results to the official test sever. The results are reported in Table~\ref{tab:server}.

\textbf{Results} From Table~\ref{tab:state} and Table~\ref{tab:server}, we can see that in most cases, SCA-CNN outperforms the other models. This is due to
the fact that SCA-CNN exploits spatial, channel-wise, and multi-layer attentions, while most of other attention models only consider one attention type. The
reasons why we cannot surpass ATT and Google NIC come from two sides: 1) Both ATT and Google NIC use ensemble models, while SCA-CNN is a single model;
ensemble models can always obtain better results than single one. 2) More advanced CNN architectures are used; as Google NIC adopts
Inception-v3~\cite{szegedy2016rethinking} which has a better classification performance than ResNet which we adopted. In local experiments, on the MSCOCO
dataset, ATT surpasses SCA-CNN only $0.6\%$ in BLEU4 and $0.1\%$ in METEOR, respectively. For the MSCOCO server results, Google NIC surpass SCA-CNN only
$0.7\%$ in BLEU4 and $1\%$ in METEOR, respectively.

\begin{figure*}[htbp]
	\centering
	\includegraphics[width=.95\linewidth]{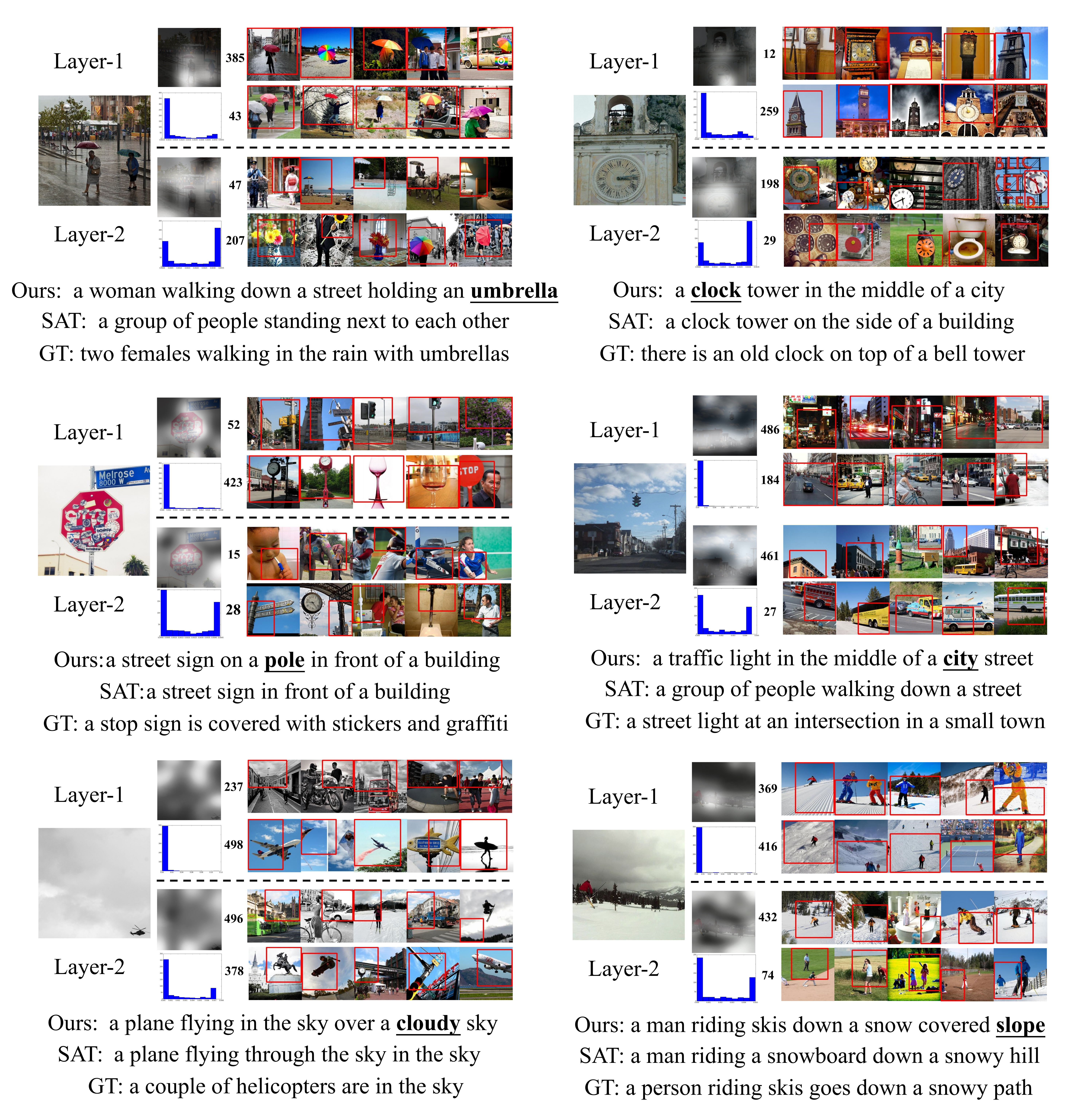}
	\caption{Examples of visualization results on spatial attention and channel-wise attention. Each example contains three captions. Ours(SCA-CNN), SAT(hard-attention) and GT(ground truth). The numbers in the third column are the channel numbers of VGG-19 network with highest channel attention weights, and next five images are selected from MSCOCO train set with high activation in the corresponding channel. The red boxes are respective fields in their corresponding layers}
	\label{fig:3}
\end{figure*}

\subsection{Visualization of Spatial and Channel-wise Attention}
We provided some qualitative examples in Figure~\ref{fig:3} for a better understanding of our model. For simplicity, we only visualized results at one word
prediction step. For example in the first sample, when SCA-CNN model tries to predict word \texttt{umbrella}, our channel-wise attention will assign more
weights on feature map channels generated by filters according to the semantics like umbrella, stick, and round-like shape. The histogram in each layer
indicates the probability distribution of all channels. The map above histogram is the spatial attention map and white indicates the spatial regions where the
model roughly attends to. For each layer we selected two channels with highest channel-wise attention probability. To show the semantic information of the
corresponding CNN filter, we used the same methods in~\cite{zeiler2014visualizing}. And the red boxes indicate their respective fields.

%% file: conc.tex
\section{Conclusions}
In this paper, we proposed a novel deep attention model dubbed SCA-CNN for image captioning. SCA-CNN takes full advantage of characteristics of CNN to yield
attentive image features: spatial, channel-wise, and multi-layer, thus achieving state-of-the-art performance on popular benchmarks. The contribution of
SCA-CNN is not only the more powerful attention model, but also a better understanding of where (\ie, spatial) and what (\ie, channel-wise) the attention
looks like in a CNN that evolves during sentence generation. In future work, we intend to bring temporal attention in SCA-CNN, in order to attend features in
different video frames for video captioning. We will also investigate how to increase the number of attentive layers without overfitting.